\documentclass[lettersize,journal]{IEEEtran}

\usepackage{amsmath,amsfonts}
\usepackage{algorithmic}
\usepackage{array}
\usepackage[caption=false,font=normalsize,labelfont=sf,textfont=sf]{subfig}
\usepackage{textcomp}
\usepackage{stfloats}
\usepackage{url}
\usepackage{verbatim}
\usepackage{graphicx}

\usepackage{hyperref}
\usepackage{xcolor,colortbl}
\usepackage{multirow}
\usepackage{booktabs}
\usepackage{makecell}
\usepackage{color}
\newcolumntype{M}[1]{>{\centering\arraybackslash}m{#1}}

\def\BibTeX{{\rm B\kern-.05em{\sc i\kern-.025em b}\kern-.08em
    T\kern-.1667em\lower.7ex\hbox{E}\kern-.125emX}}
\usepackage{balance}
\begin{document}
\title{Remote Sensing Large Vision-Language Model: Semantic-augmented Multi-level Alignment and Semantic-aware Expert Modeling}
\author{Sungjune Park*, Yeongyun Kim*, Se Yeon Kim*, and Yong Man Ro$^\dagger$,~\IEEEmembership{Senior Member,~IEEE}%
\thanks{S. Park, Y. Kim, S. Y. Kim, and Y. M. Ro are with Integrated Vision and Language Lab., School of Electrical Engineering, Korea Advanced Institute of Science and Technology (KAIST), 291 Daehak-ro, Yuseong-gu, Daejeon, 34141, Republic of Korea (e-mail: sungjune-p@kaist.ac.kr; yeongyun.kim@kaist.ac.kr; seyeon.kim@kaist.ac.kr; ymro@kaist.ac.kr). \\
* Equal contribution. \quad $^\dagger$ Corresponding author: Y. M. Ro. (ymro@kaist.ac.kr)}}

\maketitle
\begin{abstract}
    Large Vision and Language Models (LVLMs) have shown strong performance across various vision-language tasks in natural image domains. However, their application to remote sensing (RS) remains underexplored due to significant domain differences in visual appearances, object scales, and semantics. These discrepancies hider the effective understanding of RS scenes, which contain rich, multi-level semantic information spanning from coarse-to-fine levels. Hence, it limits the direct adaptation of existing LVLMs to RS imagery. To address this gap, we propose a novel LVLM framework tailored for RS understanding, incorporating two core components: Semantic-augmented Multi-level Alignment and Semantic-aware Expert Modeling. First, to align multi-level visual features, we introduce the retrieval-based Semantic Augmentation Module which enriches the visual features with relevant semantics across fine-to-coarse levels (\textit{e.g.,} object- and scene-level information). It is designed to retrieve relevant semantic cues from a RS semantic knowledge database, followed by aggregation of semantic cues with user query and multi-level visual features, resulting in semantically enriched representation across multiple levels. Second, for Semantic-aware Expert Modeling, we design semantic experts, where each expert is responsible for processing semantic representation at different levels separately. This enables hierarchical semantic understanding from coarse to fine levels. Evaluations across multiple RS tasks—including scene classification and VQA \textit{etc.}—demonstrate that the proposed framework achieves consistent improvements across multiple semantic levels. This highlights its capability and effectiveness in bridging the gap between general LVLMs and unique demands of RS-specific vision-language understanding.
\end{abstract}

\section{Introduction}
\IEEEPARstart{T}{hese} days, the rapid advancements in Large Language Models (LLMs)~\cite{llama, vicuna, phi3, alpaca, gpt4} and Vision-Language Pretraining (VLP)~\cite{clip, blip2}, along with the emergence of numerous instruction-tuning datasets~\cite{instructblip, llava}, have led to remarkable accomplishment in Large Vision and Language Models (LVLMs)~\cite{cogvlm, mplug-owl2, flamingo, internvl, gpt4v}. Building upon strong foundation models like LLaVA~\cite{llava}, an increasing number of recent studies have been introduced to further push the boundaries of LVLMs, exploring their capabilities, potential, and limitations across diverse scenarios. These LVLMs have demonstrated impressive performance across a wide range of vision-language tasks, including image captioning, visual question answering (VQA), visual grounding, and instruction-following tasks which demand complex reasoning, spatial understanding, and multi-step decision making~\cite{qwen2.5-vl, cot, deepseek-vl2, visionllm, vision-r1}. Their success has proven the potential of LVLMs to serve as general-purpose vision-language agents, capable of interpreting, reasoning over, and interacting with both visual and textual modalities in a human-like manner.

Meanwhile, remote sensing (RS) research has gained increasing importance with the increasing availability of high-resolution satellite imagery and the growing demand for automated geospatial analysis in areas such as climate monitoring, urban planning, and disaster response~\cite{rs1, rs2, rs3, rs4, rs5, rs6, rs7}. However, despite the remarkable progress of LVLMs in general-purpose domain, the development of LVLMs in the RS domain remains significantly underexplored~\cite{geochat, lhrs-bot, skyeyegpt}. It is primarily caused by the substantial domain discrepancy between natural and RS images, making it challenging to directly apply generic LVLMs to RS domain. To address such a problem, several recent works have emerged to adapt LVLMs to RS applications by bridging the domain gap (\textit{e.g.,} domain discrepancies in visual patterns, object scales, and semantic distributions)~\cite{h2rsvlm, rs-llava, skysense}. For example, RSGPT~\cite{rsgpt} concentrates on aligning RS images with instruction-tuned textual prompts from RSICap. GeoChat~\cite{geochat} demonstrates multitask capabilities using high-resolution RS images. RSUniVLM~\cite{rsunivlm} designs granularity experts, which is task-specific encoding experts depending on input data type. LHRS-Bot~\cite{lhrs-bot} employs a vision perceiver leveraging multi-level visual features to improve vision-language alignment. While these approaches have shown impressive progress in RS LVLMs, existing methods still fall short in effectively modeling multi-level, hierarchical semantics which is inherent in RS imagery. A key aspect in understanding RS imagery lies in its inherent characteristic, diverse semantic levels within a single RS image. A single RS image contains both coarse-grained semantics (\textit{e.g.,} land usage purpose, urban/rural scenes) and fine-grained details (\textit{e.g.,} objects' types, textures, spatial layouts). To effectively interpret these diverse semantic cues, LVLM is expected to understand and reason across different semantic levels. This requires not only broad coarse-grained level semantic comprehension but also accurate interpretation of fine-grained level contents. Therefore, it is desirable to incorporate hierarchical modeling, enabling LVLMs to effectively understand semantics across varying levels of granularity. However, most existing methods lack the ability to handle semantics in a hierarchical manner and instead process all semantic information simultaneously.

\begin{figure*}[ht!]
    \centering
    \includegraphics[width=0.99\textwidth]{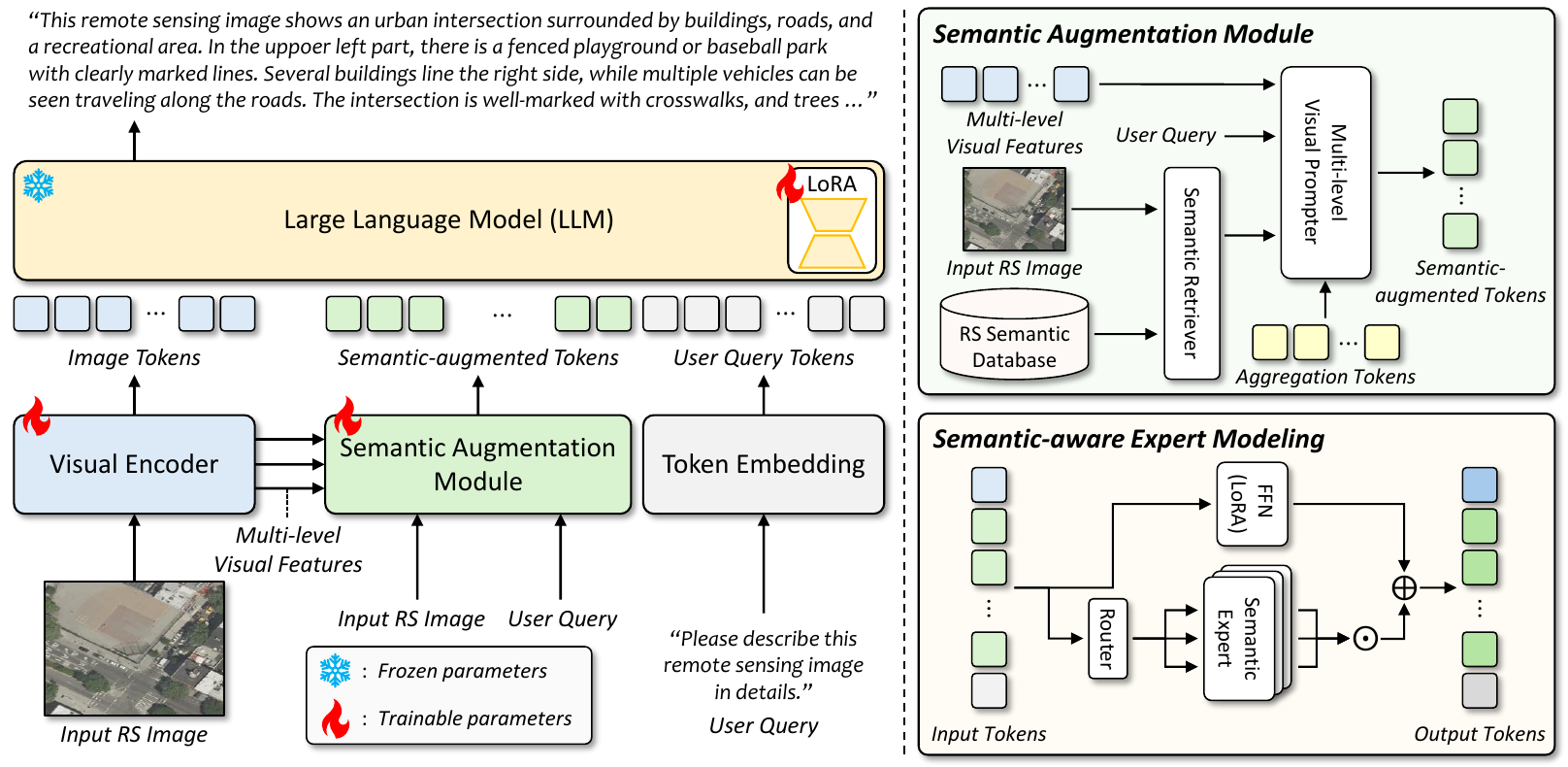}
    \vspace{-0.1cm}
	\caption{Overview of the proposed framework. The Semantic Augmentation Module retrieves the descriptions, most relevant to the given RS image using a CLIP-based semantic retriever. Those descriptions are then integrated with the user query and multi-level visual features (obtained from the visual encoder) within the multi-level visual prompter utilizing learnable aggregation tokens. The resulting multi-level semantic-augmented visual prompt tokens (\textit{green tokens}) are passed to the LLM along with the image tokens (\textit{blue tokens}) and the user query tokens (\textit{gray tokens}). For Semantic-aware Expert Modeling, $L$ semantic experts are placed, and a router assigns each level of semantic visual representation to the corresponding expert, enabling the LLM to encoder and reason over hierarchical semantics in a disentangled manner.}
	\label{fig1}
    \vspace{-0.3cm}
\end{figure*}

In this paper, we propose a novel RS LVLM framework, which aims to enhance multi-level semantic understanding of RS imagery from scene-level contexts to object-level details. Our framework, illustrated in Fig.~\ref{fig1}, is designed to explicitly incorporate hierarchical semantic cues, enabling LVLMs to understand and reason across fine-to-coarse semantic levels. Specifically, we introduce two core components: Semantic-augmented Multi-level Alignment and Semantic-aware Expert Modeling. Semantic-augmented Multi-level Alignment is a retrieval-based semantic enriching module, composed of three submodules: (1) a RS semantic knowledge database, (2) a CLIP-based semantic retriever, and (3) a multi-level visual prompter. The database contains millions of textual descriptions illustrating RS scenes across multiple semantic levels, from scene-level contexts to object-level details. Given an RS image, the retriever fetches the descriptions that are most semantically relevant to the visual contents. Next, the multi-level visual prompter aligns visual and textual semantics across different levels, by integrating the semantic information from the retrieved descriptions with both user query and multi-level visual features. This process yields semantically enriched visual prompts which condition the LLM's responses with rich semantic contexts in diverse levels. To handle hierarchical semantics effectively within the LLM, we devise Semantic-aware Expert Modeling. By placing semantic experts, different levels of semantics are processed separately. The visual prompts, corresponding to different semantic levels, are routed to each semantic expert, where each expert is specialized in processing specific semantic granularity. This expert-based routing enables the LLM to learn disentangled semantic representations and effectively understand hierarchical, coarse-to-fine contexts in RS imagery. The proposed framework is trained with a two-stage procedure and evaluated on multiple RS vision-language benchmarks, including scene classification, visual question answering (VQA), visual grounding, and image captioning. Experimental results demonstrate the effectiveness of the proposed method across diverse tasks and semantic levels, corroborating its comprehensive RS understanding.

The main contribution of the proposed method can be summarized as follows:
\begin{itemize}
    \item We present a novel LVLM framework for the RS domain incorporating Semantic-augmented Multi-level Alignment and Semantic-aware Expert Modeling to enhance multi-level semantic understanding.   
    \item We introduce a retrieval-based Semantic Augmentation Module, which leverages a curated RS semantic database to enrich visual prompts with relevant multi-level textual descriptions, ranging from coarse to fine semantics.
    \item We devise Semantic-aware Expert Modeling, in which level-specific semantic experts are assigned to separately process coarse- and fine-grained visual semantics.    
    \item Experimental results across diverse RS vision-language tasks demonstrate the effectiveness of the proposed framework, showing consistent performance improvements over existing baselines.
\end{itemize}

\section{Related Work}
\begin{table*}[ht!]
	\renewcommand{\arraystretch}{1.3}
	\renewcommand{\tabcolsep}{5.0mm}
    \centering
    \caption{Examples of descriptions stored in the RS semantic knowledge database, along with the corresponding reference images. Note that these images are not stored in the database, but included to help visualize the associated descriptions.}
    \resizebox{0.99\linewidth}{!}{
    \begin{tabular}{cM{11cm}}
    \toprule[1.3pt]
    \bf RS Image & \bf Descriptions \\
    \midrule
    \raisebox{-.45\height}{\includegraphics[width=0.13\linewidth]{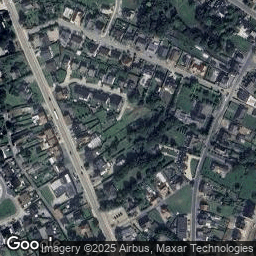}} & The image presents a detailed aerial view of \textbf{a residential area}. The landuse is predominantly residential, characterized by the presence of \textbf{numerous houses} with varying roof colors, primarily in shades of brown and gray. \textbf{The greenfield areas} are interspersed between the residential zones, providing a natural contrast to the built environment. These green spaces are likely parks or open fields, contributing to the livability of the area. \textbf{The roads} are a network of streets that meander through the neighborhood, with some appearing to be main \textbf{thoroughfares} while others are likely smaller, \textbf{local roads}. The layout suggests \textbf{a planned community} with a mix of open spaces and residential blocks. $\dots$ \\
    \midrule
    \raisebox{-.49\height}{\includegraphics[width=0.13\linewidth]{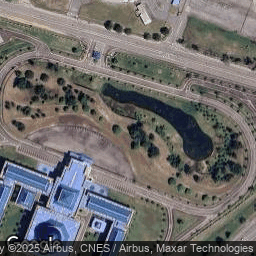}} & The image presents a detailed aerial view of \textbf{a landscaped area}, likely part of a larger urban or suburban environment. \textbf{The grassy areas} are well-maintained, with varying shades of green, indicating different types of grass or the effects of sunlight. \textbf{The water body, a pond,} is a significant feature, providing a natural element within the otherwise man-made landscape. \textbf{The asphalt surface of the highway and service road} suggests a well-connected area, possibly with a focus on accessibility and traffic flow. \textbf{The presence of wood}, likely in the form of trees or shrubs, adds to the biodiversity of the area. $\dots$ \\
    \midrule
    \raisebox{-.45\height}{\includegraphics[width=0.13\linewidth]{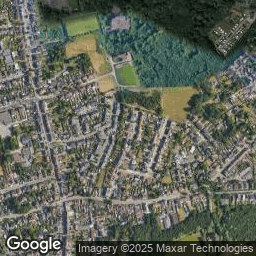}} & The image presents a detailed aerial view of \textbf{a suburban area}. The residential zones are characterized by \textbf{a dense arrangement of houses}. $\dots$ The land use is predominantly \textbf{residential}, with the \textbf{forested areas} providing a natural boundary and possibly serving as a nature reserve or park. \textbf{These green spaces} are interspersed with meadows, which could be used for recreational purposes or as part of the local ecosystem. \textbf{The trees} within the residential areas likely serve as a buffer zone between the homes and the forested land, enhancing the aesthetic appeal and providing shade. $\dots$ The overall impression is one of \textbf{a well-planned community} with a balance between urban development and natural spaces. $\dots$ \\
    \midrule
    \raisebox{-.47\height}{\includegraphics[width=0.13\linewidth]{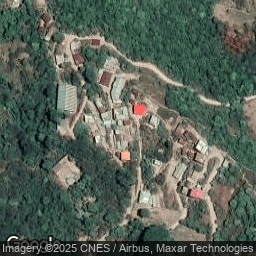}} & The image depicts \textbf{a rural village with a residential land use}. The village is characterized by \textbf{a dense cluster of buildings with varying roof colors}, predominantly red and gray. \textbf{The layout is somewhat irregular}, with buildings closely packed together, suggesting a community-oriented design. The presence of greenery indicates that the village is \textbf{surrounded by or interspersed with trees}, which could provide a natural barrier or aesthetic appeal. \textbf{The roads} are narrow and winding, typical of rural areas, and they appear to be unpaved. $\dots$ The overall impression is of \textbf{a small, close-knit community} living in a rural environment. $\dots$ \\
    \bottomrule[1.3pt]
    \end{tabular}}
	\label{tab1}
\end{table*}

\subsection{Large Vision and Language Models (LVLMs)}
As Large Language Models (LLMs) have proven their remarkable capabilities in a wide range of language processing tasks~\cite{vicuna, llama, llama2, llama3, phi3, qwen, alpaca, internlm, gpt4}, Large Vision and Language Models (LVLMs) have emerged demonstrating successful applications of LLMs into visual domains~\cite{llava, llavanext, internvl, internvl3, flamingo, minigpt, gpt4v, mini-gemini, gemini, kosmos-2.5}. Furthermore, the development of Vision and Language Pretraining (VLP)~\cite{clip, align, blip2, cogvlm} and the emergence of numerous instruction-tuning datasets~\cite{sharegpt4v, minigpt-v2, internlm-xcomposer, gpt4roi, glamm} have led to remarkable accomplishments in aligning visual and language data, showing noticeable capabilities across a variety of vision and language tasks, such as image captioning, visual question answering (VQA), visual grounding, region captioning, object counting, and so on. For instance, LLaVA~\cite{llava} integrates a pretrained CLIP visual encoder~\cite{clip} with a language model via projection layers and is instruction-tuned for multi-turn reasoning. InstructBLIP~\cite{instructblip} extends BLIP-2~\cite{blip2} by instruction-tuning, enhancing its ability to follow detailed visual instructions and answer diverse image-based queries. InternVL~\cite{internvl} focuses on fine-grained vision-language alignment by incorporating dense region-text supervision and shows powerful performance on visual grounding and referring expression tasks. Gemini~\cite{gemini} combines visual perception and language understanding in a multitask setting, supporting general image understanding optical character recognition (OCR), and so on. Qwen2.5-VL~\cite{qwen2.5-vl} supports high-resolution image understanding with fine-grained spatial reasoning capabilities. Despite these impressive advances of LVLMs in generic domains, LVLMs in remote sensing (RS) domain remain significantly underexplored. The key aspects of RS imagery—such as domain-specific visual patterns, scale variations, and hierarchical semantic levels (different from generic domain)—pose significant challenges to direct adaption of LVLMs into the RS domain.

\subsection{Remote Sensing LVLMs}
In response to the growing demand for vision-language models tailored to remote sensing (RS), several recent studies have explored how to adapt and specialize LVLMs for the RS domain~\cite{skyeyegpt, lhrs-bot-nova, h2rsvlm, rs-llava, skysense, earthmarker, geollava, teochat}. For instance, RSGPT~\cite{rsgpt} is a generative pretrained model finetuned on curated RSICap dataset, to understand general knowledge in RS imagery. GeoChat~\cite{geochat} serves as a strong RS foundation LVLM, trained on multimodal instruction-following datasets specific to remote sensing. EarthGPT~\cite{earthgpt} attempts to bridge the domain gap between natural and RS images by adopting a visual-enhanced perception module that combines CNN and ViT-based encoders. LHRS-Bot~\cite{lhrs-bot} employs a visual perceiver to align multi-scale visual features. RSUniVLM~\cite{rsunivlm} leverages a granularity-oriented mixture-of-experts (G-MoE) to encode pixel-, region-, and image-level features depending on RS tasks. RingMoGPT~\cite{ringmogpt} focuses on object-level recognition by incorporating a localization decoder and a Q-Former module between the image encoder and the LLM. In contrast to these approaches, our proposed method aims to equip LVLMs with a deeper understanding of hierarchical semantic information in RS imagery through two key mechanisms: \textbf{Semantic-augmented Multi-level Alignment} and \textbf{Semantic-aware Expert Modeling}. To this end, we construct a retrieval-based semantic augmentation module that retrieves detailed textual descriptions relevant to the input RS image, and inject these semantics while aligning multi-level visual features. Furthermore, we design a set of semantic experts within the LLM to encode different semantic levels separately, enabling comprehensive visual understanding within a unified framework.

\section{Methodology}
In this section, we elaborate on the details of the proposed framework which consists of two core components: 1) \textbf{Semantic-Augmented Multi-level Alignment}, composed of three submodules--a RS semantic knowledge database, a semantic retriever, and a multi-level visual prompter, and 2) \textbf{Semantic-Aware Expert Modeling}, which is designed to capture multi-level semantics via hierarchical semantic experts. The overview of our framework is illustrated in Fig.~\ref{fig1}.

\subsection{Semantic-augmented Multi-level Alignment}
Semantic-augmented Multi-level Alignment aims to obtain multi-level semantic-augmented tokens, semantically enriched across hierarchical (fine-to-coarse) levels, while aligning multi-level visual features. These tokens provides hierarchical semantic cues, guiding the LLM to generate responses via comprehensive semantic understanding.

\subsubsection{RS Semantic Database and Semantic Retriever}
\label{retrieve}
To enable semantic retrieval, we first construct a RS semantic knowledge database containing diverse descriptions across multiple semantic levels. The database is designed to include both coarse-grained information (\textit{e.g.,} scene category, location type) and fine-grained details (\textit{e.g.,} object types, spatial layouts). We employ the existing remote sensing caption data, LHRS-Align-Recap~\cite{lhrs-bot-nova}, which contains detailed and well-structured textual descriptions for a variety of RS scenes. The descriptions includes OpenStreetMap (OSM) semantic attributes\footnote{\url{https://wiki.openstreetmap.org/wiki/Map_features}}, encompassing the types and purposes of places, buildings, and other objects in RS images. The examples of textual descriptions are introduced in TABLE~\ref{tab1}. As shown in the table, it contains semantic information across diverse levels. For instance, the last row illustrates both high-level context such as scene type and purpose (\textit{e.g.,} a rural village with a residential function and a small, close-knit community) and fine-grained details, including object types and spatial layouts(\textit{e.g.,} a dense cluster of buildings with varying roof colors, irregular layouts, trees, and roads). Based on such observations, we posit that such textual descriptions include semantic information at multiple levels, so that it would be suitable and fit to be employed for the RS semantic database.

\begin{figure}[t!]
    \centering
    \includegraphics[width=0.999\linewidth]{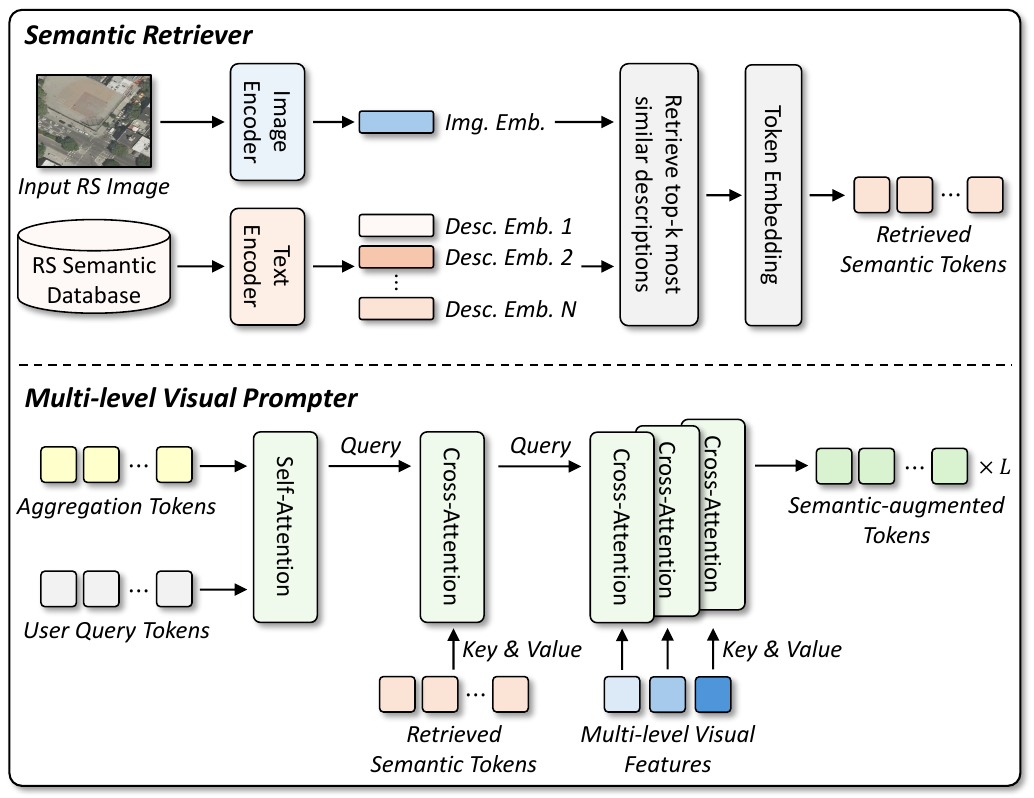}
	\caption{The upper figure shows the overall process in the semantic retriever. By comparing image and textual description embeddings, it fetches most relevant $k$ descriptive semantics. Through the token embedding, it obtains the retrieved semantic tokens. The lower figure illustrates the detailed architecture of the multi-level visual prompter. By utilizing the learnable aggregation tokens, it aggregates information from the user query, retrieved semantic cues, and multi-level visual features.}
	\label{fig2}
\end{figure}

Once the RS semantic database is constructed, the next step is to retrieve the informative semantics from the database. The informative semantics correspond to the most relevant textual descriptions which properly illustrate the given RS scenes with its contextual attributes. To this end, we adopt CLIP image and text encoders as the retriever backbone, and fine-tune these encoders with paired RS image and text data (\textit{i.e.,} a subset of LHRS-Align-Recap). Through contrastive learning, the encoders are able to generate embeddings that are more representative to the RS scenes and their associated descriptions. After constructing the CLIP-based retriever, the overall process of fetching relevant descriptions is described in Fig~\ref{fig2}. Given the input RS image, the image encoder extracts an image embedding, while the text encoder extracts the description embeddings. Cosine similarity is computed between the image embedding and each of the description embeddings. Based on the similarity scores, it retrieves the top $k$ relevant descriptions, which are most semantically aligned with the given RS scene. These descriptions are subsequently converted into semantic tokens via token embedding and fed into the multi-level visual prompter (which is elaborated in the following subsection). Through this retrieval process, it provides multi-level semantics, thereby enhancing its ability to understand and reason about hierarchical attributes of RS images.

\subsubsection{Multi-level Visual Prompter}
The multi-level visual prompter is designed to integrate information from three sources: the user query--which includes user's intention, retrieved semantic descriptions--which encompass diverse semantic information, and multi-level visual features--which are extracted from the visual encoder. The goal is to obtain semantically enriched visual prompts (\textit{i.e.,} semantic-augmented tokens) corresponding to different semantic levels. Fig.~\ref{fig2} illustrates the overall architecture of the multi-level visual prompter. In summary, the overall procedure first attends to the user query tokens which contain the user's intent, followed by referencing to the retrieved semantic information. Finally, it aggregates visual contexts at each different semantic levels to construct multi-level semantic-augmented representations.

Specifically, as shown in the figure, the learnable aggregation tokens $\boldsymbol{f_{\text{agg}}} \in \mathbb{R}^{N_a \times d}$, where $N_a$ and $d$ denote the number of aggregation tokens and token dimension, are introduced to aggregate the information of the user query tokens, retrieved semantic tokens, and multi-level visual features. First, the aggregation tokens are concatenated with the user query tokens $\boldsymbol{f_{\text{user}}} \in \mathbb{R}^{N_u \times d}$, where $N_u$ is the number of user query tokens, and then the concatenated tokens are fed into self-attention block as follows:
\begin{equation}
\begin{aligned}
    \boldsymbol{f_{\text{in}}} = \;& \text{Concat}(\boldsymbol{f_{\text{agg}}}, \boldsymbol{f_{\text{user}}}) \in \mathbb{R}^{(N_a + N_u) \times d}, \\
    \boldsymbol{f_{\text{out}}} &= \text{SelfAttn}(\boldsymbol{f_{\text{in}}}) \in \mathbb{R}^{(N_a + N_u) \times d}.
\end{aligned}
\end{equation}
\noindent
Then we only use the output tokens corresponding to the aggregation tokens as follows:
\begin{equation}
    \boldsymbol{z^{(1)}_{\text{agg}}} = \boldsymbol{f_{\text{out}}}[:N_a] \in \mathbb{R}^{N_a \times d}.
\end{equation}
\noindent
These latent aggregation tokens $\boldsymbol{z^{(1)}_{\text{agg}}}$ are obtained after referring to the user query tokens, which contain user's intention. Second, $\boldsymbol{z^{(1)}_{\text{agg}}}$ is fed into the cross-attention module as query features, while the retrieved semantic tokens $\boldsymbol{f_{\text{semantic}}} \in \mathbb{R}^{N_s \times d}$ are employed as key and value features. Here, $N_s$ is the number of semantic tokens. By doing so, it allows the aggregation tokens to attend to semantically relevant information, yielding the second latent aggregation token $\boldsymbol{z^{(2)}_{\text{agg}}}$ as follows:
\begin{equation}
    \boldsymbol{z^{(2)}_{\text{agg}}} = \text{CrossAttn}(\boldsymbol{z^{(1)}_{\text{agg}}}, \boldsymbol{f_{\text{semantic}}}, \boldsymbol{f_{\text{semantic}}}) \in \mathbb{R}^{N_a \times d}.
\end{equation}
\noindent
Here, the query, key, and value projection matrices are omitted for the simplicity. So that, $\boldsymbol{z^{(2)}_{\text{agg}}}$ refer to both user query content and informative semantic information of multiple levels. Lastly, we pass $\boldsymbol{z^{(2)}_{\text{agg}}}$ through $L$ parallel cross-attention modules, where each cross-attention module is corresponding to a specific level of visual features extracted from the visual encoder. Thus, we also prepare $L$ levels of visual features from the visual encoder, $\boldsymbol{F_\text{vis}}=\{\boldsymbol{f^l_{\text{vis}}}\}^L_{l=1}$, where $\boldsymbol{f^l_{\text{vis}}} \in \mathbb{R}^{N_l \times d_l}$. Here, $N_l$ and $d_l$ are the number of patch tokens and the dimension at the $l$-th level of visual features, respectively. For each level $l$, we utilize $\boldsymbol{z^{(2)}_{\text{agg}}}$ as query features and $\boldsymbol{f^l_{\text{vis}}}$ as key and value features in the $l$-th cross-attention module:
\begin{equation}
    \boldsymbol{s^l_{\text{agg}}} = \text{CrossAttn}(\boldsymbol{z^{(2)}_{\text{agg}}}, \boldsymbol{f^l_{\text{vis}}}, \boldsymbol{f^l_{\text{vis}}}) \in \mathbb{R}^{N_a \times d}.
\end{equation}
\noindent
Then $\boldsymbol{s^l_{\text{agg}}}$ captures the visual context of the $l$-th level, after being conditioned on both the user query and semantic information. Finally, we concatenate all outputs in the level wise to obtain the final semantic-augmented tokens as follows:
\begin{equation}
    \boldsymbol{S} = \text{Concat}(\boldsymbol{s^1_{\text{agg}}}, \boldsymbol{s^2_{\text{agg}}}, \cdots, \boldsymbol{s^L_{\text{agg}}}) \in \mathbb{R}^{(N_a \times L) \times d}.
    \label{eq5}
\end{equation}
\noindent
In summary, the multi-level visual prompter places the aggregation tokens to refer to the user query content and semantic information sequentially. These tokens are then utilized to extract level-specific visual representations across multiple semantic levels. By incorporating these representations, the semantic-enriched tokens $\boldsymbol{S}$ are augmented with rich semantic cues which are aligned with both retrieved semantic information and visual contexts. Finally, as shown in Fig.~\ref{fig1}, the obtained semantic-augmented tokens are fed into the LLM, along with image patch tokens and the user query tokens.

\subsection{Semantic-aware Expert Modeling}
\label{expert}
\begin{figure}[t!]
    \centering
    \includegraphics[width=0.99\linewidth]{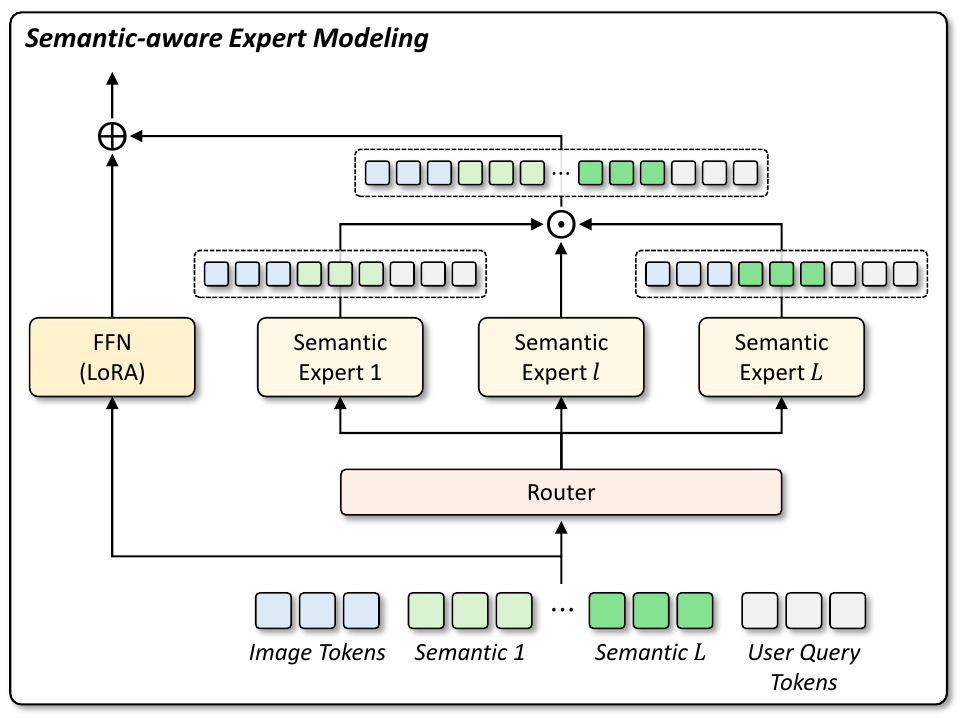}
    \vspace{-0.5cm}
	\caption{The details of Semantic-aware Expert Modeling. The input tokens includes multi-level semantic-augmented tokens along with the image and user query tokens. Given the input tokens, the router assigns each semantic expert to process semantic-augmented tokens at specific semantic level with the other image and user query tokens. Then the outputs from each expert are merged, and then integrated with the output of FFN.}
	\label{fig3}
    \vspace{-0.3cm}
\end{figure}

The LLM takes three kinds of inputs: the image patch tokens, the user query tokens, and the semantic-augmented tokens. With such input tokens, the LLM aims to generate adequate responses. Given a RS image, to fully understand the image and generate contextually appropriate response, the LLM needs to comprehend hierarchical semantics. Therefore, to enable hierarchical semantic understanding, we devise Semantic-aware Expert Modeling, integrating semantic experts where each expert processes a specific level of semantics. Fig.~\ref{fig3} illustrates the details of Semantic-aware Expert Modeling consisting of $L$ semantic experts. While the three types of input tokens are passed through the LLM, additional $L$ numbers of semantic experts are placed in parallel with the feed-forward network (FFN). When feeding the input tokens to the semantic experts, the router manipulate the input tokens such that each expert independently processes a distinct level of hierarchical semantics. When routing input tokens to the $l$-th expert, the router masks out the semantic-augmented tokens corresponding to all levels except the $l$-th level, ensuring that each expert attends only to its designated semantic level. Therefore, each semantic expert encodes the corresponding semantic representation with the image and user query tokens at the same time.

Specifically, let the entire set of input tokens be defined as follows:
\begin{equation}
    \boldsymbol{X}=[\boldsymbol{X_{\text{img}}}; \boldsymbol{X^1_{\text{S}}}; \boldsymbol{X^2_{\text{S}}}; \cdots; \boldsymbol{X^L_{\text{S}}}; \boldsymbol{X_{\text{query}}}],
\end{equation}
\noindent
where $\boldsymbol{X^l_{\text{S}}}$ denotes the input semantic-augmented intermediate tokens at the $l$-th semantic level. These tokens are corresponding to the intermediate tokens stemmed from the semantic-augmented tokens in Equation (\ref{eq5}). For each semantic expert $E^l$ for the $l$-th level, the router applies a binary mask $\boldsymbol{\textbf{M}^l}$ which helps remain the tokens to be processed together: the image tokens $\boldsymbol{X_{\text{img}}}$, the $l$-th semantic tokens $\boldsymbol{X^l_{\text{S}}}$, and the user query tokens $\boldsymbol{X_{\text{query}}}$ by masking out the rest tokens as follows:
\begin{equation}
    \boldsymbol{\textbf{M}^l_t} = 
        \begin{cases}
            1 & \text{if } \boldsymbol{x_t} \in \{\boldsymbol{X_{\text{img}}}, \boldsymbol{X^l_{\text{S}}}, \boldsymbol{X_{\text{query}}}\}, \\
            0 & \text{otherwise}.
\end{cases}
\end{equation}
\noindent
Then the input tokens for the $l$-th semantic expert are formulated as follows:
\begin{equation}
    \boldsymbol{\tilde{X}^l} = \boldsymbol{\textbf{M}^l} \cdot \boldsymbol{X}.
\end{equation}
\noindent
The $l$-th semantic expert takes $\boldsymbol{\tilde{X}^l}$ as input. Each expert is designed as a low-rank feed-forward projection architecture, consisting of a rank-reduction matrix $\boldsymbol{U^l} \in \mathbb{R}^{d_h \times d_r}$ and a rank-expansion matrix $\boldsymbol{V^l} \in \mathbb{R}^{d_r \times d_h}$. Here, $d_h$ denotes the dimension of the token hidden states, and $d_r$ is the reduced rank, where $d_r \ll d_h$. Such an architecture design allows efficient processing for each semantic level. The hidden outputs of the $l$-th expert $E^l$ are obtained as follows:
\begin{equation}
    \boldsymbol{h^l} = E^l(\boldsymbol{\tilde{X}^l}) = \boldsymbol{V^l} \cdot \boldsymbol{U^l} \cdot \boldsymbol{\tilde{X}^l}.
\end{equation}
\noindent
The outputs from every semantic expert are then merged as follows (denoted by $\odot$ in Fig.~\ref{fig3}):
\begin{equation}
    \boldsymbol{h_s} = \boldsymbol{h^1} \odot \boldsymbol{h^2} \odot \cdots \odot \boldsymbol{h^L} = \sum_{l=1}^{L} g_l\;\boldsymbol{h^l},
    \label{eq_gate}
\end{equation}
\noindent
where $g_l$ represents the gating weight for the $l$-th semantic expert outputs. It is obtained from a softmax function followed by a gating projection layer $\boldsymbol{W_g} \in \mathbb{R}^{d_h \times L}$. Finally, we incorporate the merged semantic output $\boldsymbol{h_s}$ with the output of the standard FFN by element-wise summation as follows (denoted by $\oplus$ in Fig.~\ref{fig3}):
\begin{equation}
    \boldsymbol{h} = \text{FFN}(\boldsymbol{X_l}) + \boldsymbol{h_s}.
\end{equation}
\noindent
Semantic-aware Expert Modeling enables the LLM to selectively and efficiently process visual features at multiple semantic levels, thereby supporting coarse-to-fine visual understanding within a unified framework.

\section{Experiment}
\subsection{Experimental Setup}
\subsubsection{Datasets}
For the image-text pair alignment training, we use a total of 1.8M image-caption data (300K of LHRS-Align~\cite{lhrs-bot} and 1.5M of SkyScript~\cite{skyscript}). Before directly using SkyScript caption data, Qwen2.5-14B-Instruct~\cite{qwen2.5-vl} is utilized to paraphrase the captions to be in more structured format, providing the reference sentences from LHRS-Align. Then we use 395.4K data for the instruction-tuning step with following datasets: UCM~\cite{ucm}, NWPU~\cite{nwpu, nwpu-cap}, RSITMD~\cite{rsitmd}, METER-ML~\cite{meterml}, fMoW~\cite{fmow}, RSICD~\cite{rsicd}, RSVQA~\cite{rsvqa}, RSVG~\cite{rsvg}, DIOR-RSVG~\cite{dior-rsvg}, DIOR~\cite{dior}, and DOTA~\cite{dota}. The detailed statistics of each dataset are shown in TABLE~\ref{tab2}. To verify the effectiveness of our method, the proposed framework is evaluated on six scene classification datasets; AID~\cite{aid}, WHU-RS19~\cite{whu-rs19}, NWPU~\cite{nwpu}, SIRI-WHU~\cite{siri-whu}, METER-ML~\cite{meterml}, and fMoW~\cite{fmow}, two image captioning datasets; NWPU-Captions~\cite{nwpu-cap} and UCM~\cite{ucm}, two visual question answering (VQA) datasets; RSVQA-LR and -HR~\cite{rsvqa}, and a visual grounding dataset; DIOR-RSVG~\cite{dior}.

\definecolor{Gray}{gray}{0.93}
\definecolor{Green}{rgb}{0.9, 0.95, 0.97}

\begin{table}[t!]
	\renewcommand{\arraystretch}{1.2}
	\renewcommand{\tabcolsep}{5.0mm}
    \centering
    \caption{The detailed statistics of instruction-tuning data.}
    \resizebox{0.99\linewidth}{!}{
    \begin{tabular}{lcc}
    \toprule[1.3pt]
    \bf Dataset & \bf RS Task & \bf \# Samples \\
    \midrule
    \rowcolor{Gray}
    UCM~\cite{ucm} & Scene Classification & 2K \\
    \rowcolor{Gray}
    NWPU~\cite{nwpu} & Scene Classification & 5K \\
    \rowcolor{Gray}
    RSITMD~\cite{rsitmd} & Scene Classification & 0.5K \\
    \rowcolor{Gray}
    METER-ML~\cite{meterml} & Scene Classification & 1.4K \\
    \rowcolor{Gray}
    fMoW~\cite{fmow} & Scene Classification & 5.3K \\
    RSICD~\cite{rsicd} & Image Captioning & 1K \\
    NWPU-Captions~\cite{nwpu-cap} & Image Captioning & 1K \\
    RSITMD~\cite{rsitmd} & Image Captioning & 12K \\
    \rowcolor{Gray}
    RSVQA-LR~\cite{rsvqa} & VQA & 5.5K \\
    \rowcolor{Gray}
    RSVQA-HR~\cite{rsvqa} & VQA & 106K \\
    RSVG~\cite{rsvg} & Visual Grounding & 5K \\
    DIOR-RSVG~\cite{dior-rsvg} & Visual Grounding & 14K \\
    \rowcolor{Gray}
    DIOR~\cite{dior} & Object Detection & 29.1K \\
    \rowcolor{Gray}
    DOTA~\cite{dota} & Object Detection & 81.6K \\
    \bottomrule[1.3pt]
    \end{tabular}}
	\label{tab2}
\end{table}

\begin{table*}[ht!]
	\renewcommand{\arraystretch}{1.2}
	\renewcommand{\tabcolsep}{2.5mm}
    \centering
    \caption{The scene classification performance comparison. We measure accuracy $(\%)$ for the evaluation metric. The best and second runner-up performances are highlighted as \textbf{Bold text} and \underline{Underlined text}, respectively.}
    \resizebox{0.99\linewidth}{!}{
    \begin{tabular}{lcccccc}
    \toprule[1.3pt]
    \bf Method & \bf AID~\cite{aid} & \bf WHU-RS19~\cite{whu-rs19} & \bf NWPU~\cite{nwpu} & \bf SIRI-WHU~\cite{siri-whu} & \bf METER-ML~\cite{meterml} & \bf fMoW~\cite{fmow} \\
    \midrule
    InstructBLIP~\cite{instructblip} & 29.5 & 36.8 & 34.0 & 18.2 & 14.4 & 6.7 \\
    MiniGPTv2~\cite{minigpt-v2} & 33.0 & 64.8 & 28.2 & 35.5 & 14.3 & 5.2 \\
    LLaVA-1.5~\cite{llava-1.5} & 33.1 & 54.6 & 35.0 & 17.7 & 21.7 & 11.4 \\
    mPLUG-OWL2~\cite{mplug-owl2} & 48.8 & 72.7 & 46.6 & 54.8 & 36.3 & 17.9 \\
    InternLM-XComposer~\cite{internlm-xcomposer} & 51.6 & 72.9 & 47.5 & 46.8 & 40.2 & 11.3 \\
    Qwen-VL-Chat~\cite{qwen-vl} & 55.3 & 72.3 & 42.7 & 54.6 & 38.8 & 6.9 \\
    GeoChat~\cite{geochat} & 72.0 & 82.3 & - & 62.7 & - & - \\
    RSUniVLM~\cite{rsunivlm} & 81.2 & 84.9 & \underline{86.9} & \underline{68.1} & - & - \\
    LHRS-Bot~\cite{lhrs-bot} & \bf 91.3 & \underline{93.2} & 83.9 & 62.7 & \underline{69.8} & \bf 56.6 \\ 
    \midrule
    \bf Ours & \underline{90.8} & \bf 96.7 & \bf 90.4 & \bf 71.5 & \bf 73.6 & \underline{54.6} \\
    \bottomrule[1.3pt]
    \end{tabular}}
	\label{tab3}
\end{table*}


\begin{figure*}[ht!]
    \centering
    \includegraphics[width=0.99\linewidth]{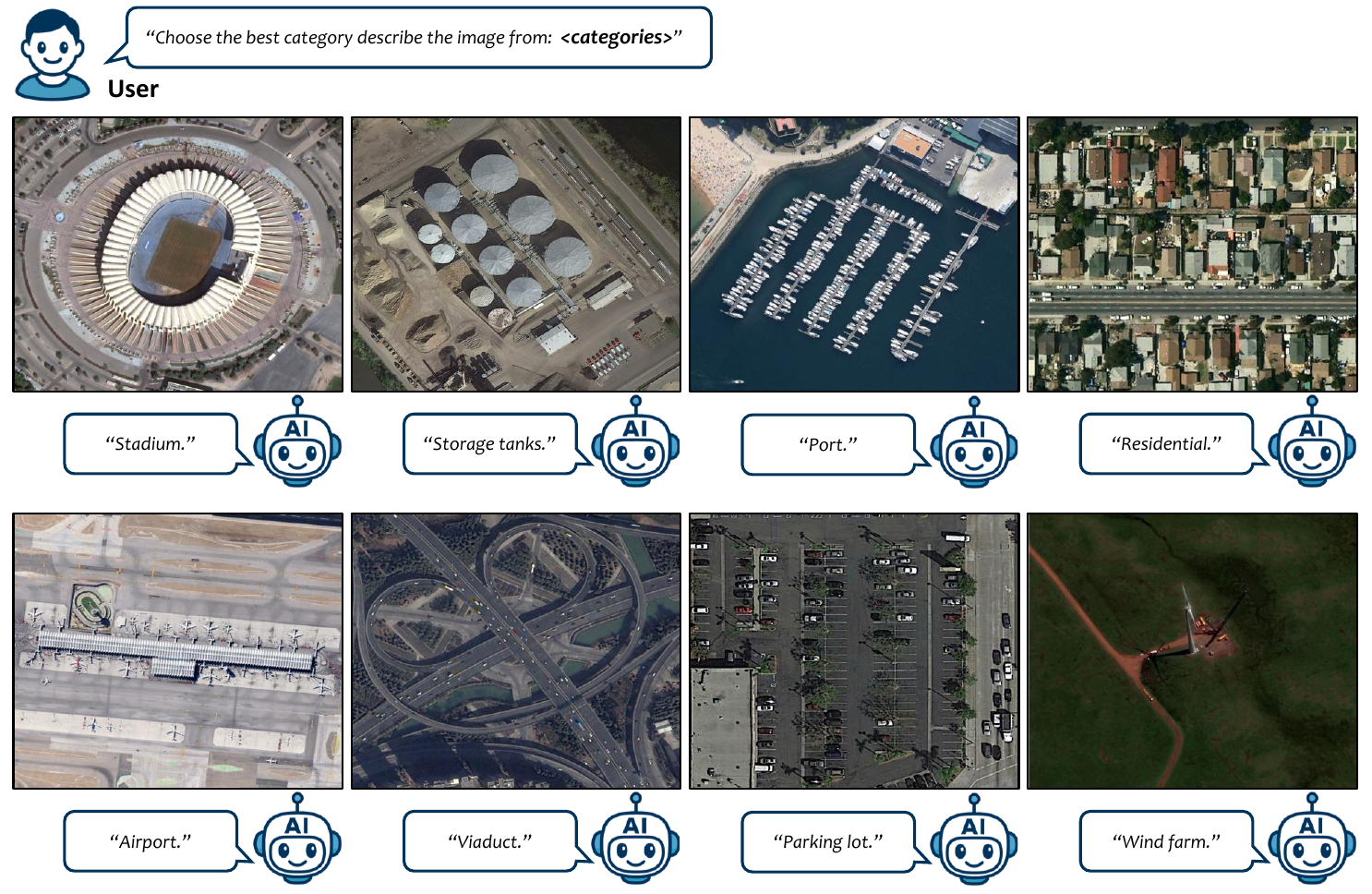}
	\caption{The qualitative visualization results of remote sensing scene classification obtained from the proposed framework.}
	\label{fig4}
\end{figure*}

\subsubsection{Evaluation Metrics}
To validate the effectiveness of the proposed method, we adopt evaluation metrics for each RS task type. For the scene classification and VQA, accuracy is adopted as the evaluation metric. For the image captioning, we use Bilingual Evaluation Understudy with 1-gram (BLEU-1), Unigram Recall-Oriented Understudy for Gisting Evaluation (ROUGE-1), Metric for Evaluation of Translation With Explicit Ordering (METEOR), and BERT F1 score. For the visual grounding, precision with intersection-over-union (IoU) threshold of $0.5$ is is adopted.

\subsubsection{Implementation Details}
The proposed framework is built upon SigLIP2~\cite{siglip2} as the visual encoder and Qwen2.5-7B-Instruct~\cite{qwen2.5} as the LLM. To capture hierarchical visual semantics, we extract the visual features from three different layers of the visual encoder located at the $1/3$, $2/3$, and last layers. Accordingly, Semantic-aware Expert Modeling incorporates three semantic experts ($L=3$), each aligned with a distinct semantic level. During the semantic retrieval, top-$5$ descriptions most relevant to the given RS images are retrieved from the database ($k=5$). We use $144$ learnable aggregation tokens, resulting in a total of $432$ semantic-augmented visual prompt tokens $(=144 \times 3)$. In Equation (\ref{eq_gate}), the gating coefficient $g_l$ is only applied to the image and user query tokens, $\boldsymbol{X_{\text{img}}}$ and $\boldsymbol{X_{\text{query}}}$. For the semantic-augmented tokens $\boldsymbol{X^l_{\text{S}}}$ corresponding to each level $l$, the coefficient $g_l$ set as $1$. The semantic experts are integrated into every fourth layer of the LLM. While the hidden size of the LLM is set to $d_h=3584$, the row rank of each semantic expert is configured as $d_r=512$. We train the proposed framework using a two-step training scheme consisting of: image-text alignment and instruction-tuning. In the first step, the framework is trained on image captioning data to align visual and textual representations. The second training step fine-tunes the framework with instruction tuning using a variety of RS tasks, such as scene classification, visual question answering (VQA), and \textit{etc.}). For both steps, standard auto-regressive language modeling objective is adopted, where the model is trained to predict the next token given the previous context. Specifically, in the first training step (\textit{i.e.,} image-text alignment), the multi-level visual prompter is trained by AdamW optimizer with a learning rate of $1e-5$ for $1$ epoch. We use $6$ NVIDIA RTX A6000 GPUs with $6$ batch sizes per each device and $4$ accumulation steps. In the second training step (\textit{i.e.,} instruction tuning), the framework is fine-tuned with following learning rates based on AdamW optimizer: $1e-6$ for the visual encoder and $1e-5$ for the multi-level visual prompter and the LLM. To efficiently adopt the LLM, we apply LoRA fine-tuning with a rank of $128$, a scaling factor of $256$, and a dropout rate of $0.05$. For this second training step, $2$ batch sizes per each device and $10$ accumulation steps are used.

\begin{table*}[ht!]
	\renewcommand{\arraystretch}{1.2}
	\renewcommand{\tabcolsep}{4.0mm}
    \centering
    \caption{The visual question answering (VQA) performance comparison on RSVQA-LR an RSVQA-HR~\cite{rsvqa}. The best and second runner-up performances are highlighted as \textbf{Bold text} and \underline{Underlined text}, respectively.}
    \resizebox{0.99\linewidth}{!}{
    \begin{tabular}{l|cccc|ccc}
    \toprule[1.3pt]
    \multirow{2}{*}{\bf Method} & \multicolumn{4}{c|}{\bf RSVQA-LR} & \multicolumn{3}{c}{\bf RSVQA-HR} \\
    \cmidrule{2-5} \cmidrule{6-8}
    & \bf Rural/Urban & \bf Presence & \bf Compare & \bf Avg. & \bf Presence & \bf Compare & \bf Avg. \\
    \midrule
    mPLUG-Owl2~\cite{mplug-owl2} & 58.0 & 74.0 & 63.7 & 65.2 & 47.6 & 58.5 & 53.0 \\
    InternLM-XCompose~\cite{internlm-xcomposer} & 59.0 & 66.7 & 52.9 & 59.6 & 67.8 & 66.6 & 67.2 \\
    LLaVA-1.5~\cite{llava-1.5} & 59.2 & 73.2 & 65.2 & 65.9 & 49.0 & 59.0 & 54.0 \\
    MiniGPTv2~\cite{minigpt-v2} & 60.0 & 51.6 & 67.6 & 59.8 & 68.3 & 64.7 & 66.5 \\
    Qwen-VL-Chat~\cite{qwen-vl} & 62.0 & 47.7 & 66.5 & 58.7 & 61.8 & 66.0 & 63.9 \\
    InstructBLIP~\cite{instructblip} & 62.6 & 48.8 & 65.9 & 59.1 & 62.6 & 62.9 & 62.8 \\
    GeoChat~\cite{geochat} & \bf 94.0 & \bf 91.1 & \underline{90.3} &\bf 90.7 & 58.5 & 83.2 & 70.8 \\
    EarthGPT~\cite{earthgpt} & - & - & - & - & 62.8 & 79.5 & 71.2 \\
    SkyEyeGPT~\cite{skyeyegpt} & 88.9 & 88.6 & 75.0 & 84.2 & 80.0 & 80.1 & 82.6 \\
    LHRS-Bot~\cite{lhrs-bot} & 89.1 & 88.5 & 90.0 & 89.2 & \underline{92.6} & \underline{92.5} & \underline{92.6} \\
    VHM~\cite{vhm} & 88.0 & 90.1 & 89.9 & 89.3 & 64.0 & 83.5 & 73.8 \\ 
    \midrule
    \bf Ours & \underline{90.0} & \underline{90.2} & \bf 90.4 & \underline{90.3} & \bf 93.1 & \bf 93.1 & \bf 93.1 \\
    \bottomrule[1.3pt]
    \end{tabular}}
	\label{tab4}
\end{table*}

\begin{figure*}[t!]
    \centering
    \includegraphics[width=0.99\linewidth]{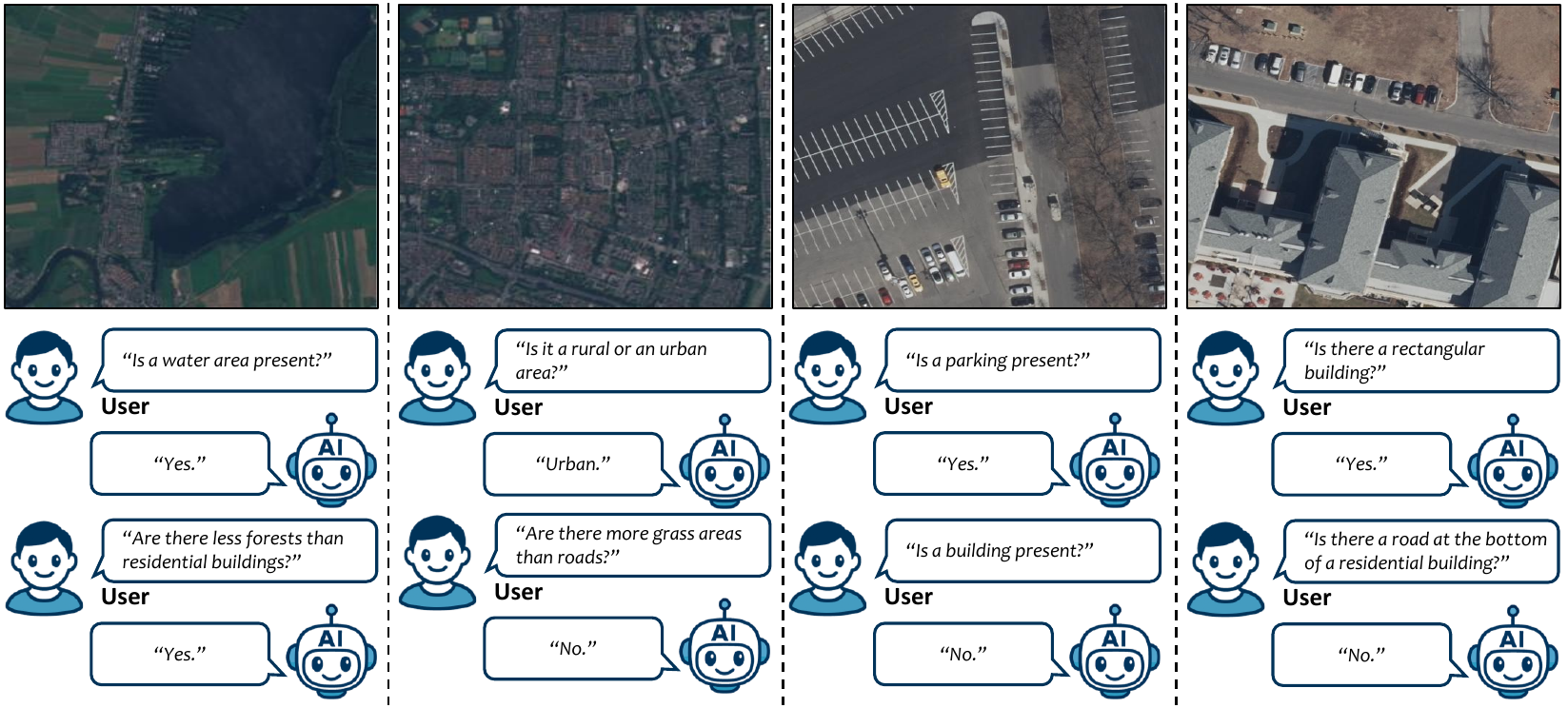}
	\caption{The qualitative visualization results of remote sensing visual question answering (VQA) obtained from the proposed framework.}
	\label{fig5}
\end{figure*}

\begin{figure*}[t!]
    \centering
    \includegraphics[width=0.99\linewidth]{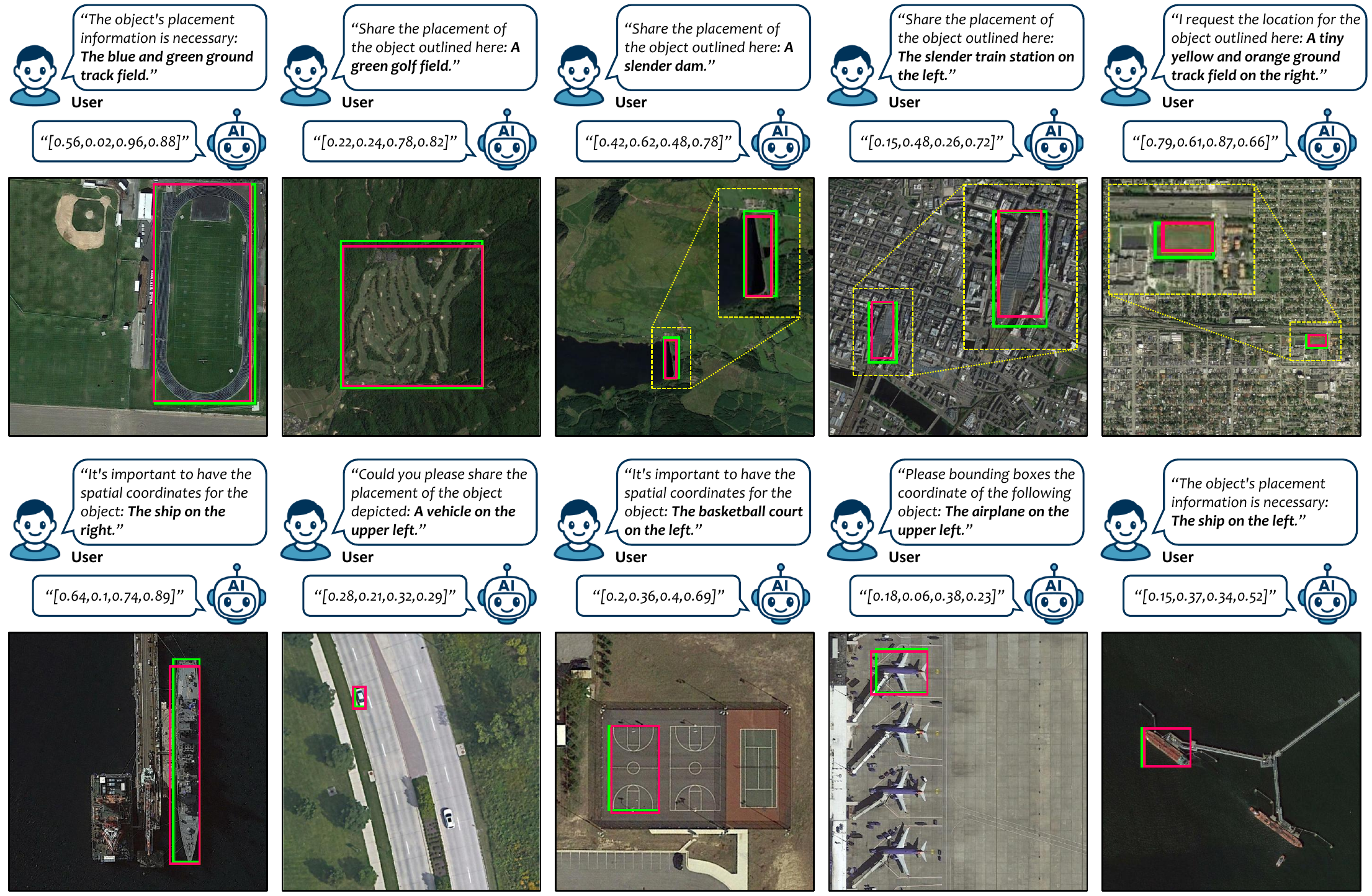}
    \vspace{-0.2cm}
	\caption{The qualitative visualization results of remote sensing visual grounding obtained from the proposed framework. The red and green boxes denote the prediction and ground-truth bounding boxes, respectively. The yellow dotted boxes expand the corresponding regions for the better visualization.}
    \vspace{-0.3cm}
	\label{fig6}
\end{figure*}

\subsection{Results on Scene Classification}
\subsubsection{Quantitative Performance Comparison}
We conduct a comprehensive evaluation on six remote sensing scene classification benchmarks, including AID~\cite{aid}, WHU-RS19~\cite{whu-rs19}, NWPU~\cite{nwpu}, SIRI-WHU~\cite{siri-whu}, METER-ML~\cite{meterml}, and fMoW~\cite{fmow}. TABLE~\ref{tab3} shows the performance comparison results. The evaluation metric is accuracy (\%). As described in the table, our proposed framework outperforms existing vision and language models across most benchmarks. Specifically, our method achieves the best performance on four out of six datasets with large margins and the second-best performance on two out of six datasets. With the first four benchmarks (AID, WHU-RS19, NWPU, and SIRI-WHU), the averaged classification accuracies of RSUniVLM~\cite{rsunivlm} and LHRS-Bot~\cite{lhrs-bot} are 80.3\% and 82.8\% respectively, while our framework obtains 87.4\% average accuracy. These experimental results on remote sensing scene classification validate the effectiveness of our hierarchical semantic understanding modeling, particularly in coarse-grained scene-level semantics, which enables better generalization across diverse scene types.

\subsubsection{Qualitative Visualization Results}
Fig.~\ref{fig4} illustrates the qualitative visualization results of remote sensing scene classification task. The input instruction, ``\texttt{Choose the best category describe the image from: <categories>}'' along with a remote sensing image, our framework responses with proper answer. Here, \texttt{<categories>} denotes the category list. For instance, we provide 20 scene categories, ``\texttt{airport, bare land, $\dots$, storage tanks, and viaduct}'', and let the model choose an answer from them. As described in the figure, the proposed framework provides proper answers across various scene types, such as stadium, storage tanks, parking lot, wind farm, and so on.

\subsection{Results on Visual Question Answering}
\subsubsection{Quantitative Performance Comparison}
The effectiveness of the proposed framework is further validated on remote sensing visual question answering (VQA) task. The comparison result is shown in TABLE~\ref{tab4}, where accuracy is used as the evaluation metric. Following the previous works~\cite{geochat, lhrs-bot, vhm}, VQA performances are reported for each question subset. As shown in the table, our framework consistently obtains robust performances on both RSVQA-LR and RSVQA-HR, where each consists of low-resolution and high-resolution remote sensing images, respectively.

\subsubsection{Qualitative Visualization Results}
Fig.~\ref{fig5} describes the examples of the qualitative visualization results on remote sensing VQA (RSVQA-LR and HR~\cite{rsvqa}), obtained from the proposed framework. As shown in the figure, when various questions are given, the proposed framework generate proper answers. For instance, when the question is about scene types: ``\texttt{Is it a rural or an urban area?}'', our framework obtains the answer ``\texttt{Urban.}'' properly. Also, when the questions ask about presence and comparison: ``\texttt{Is there a road at the bottom of a residential building?}'' and ``\texttt{Are there more grass areas than roads?}'', our model adequately generates responses by answering ``\texttt{No.}'' for both questions. As described in the visualization results, our framework, built upon hierarchical semantic understanding, can provide proper answers across diverse question types and scenes.

\begin{table}[t!]
	\renewcommand{\arraystretch}{1.2}
	\renewcommand{\tabcolsep}{7.0mm}
    \centering
    \caption{The visual grounding performance comparison on DIOR-RSVG~\cite{dior-rsvg}. The evaluation metric is the precision with 0.5 IoU threshold.}
    \resizebox{0.8\linewidth}{!}{
    \begin{tabular}{lc}
    \toprule[1.3pt]
    \bf Method & \bf Precision@0.5 \\
    \midrule
    Qwen-VL-Chat~\cite{qwen-vl} & 31.9 \\
    CogVLM~\cite{cogvlm} & 44.6 \\
    GeoChat~\cite{geochat} & 24.1 \\
    LHRS-Bot~\cite{lhrs-bot} & 19.6 \\
    VHM~\cite{vhm} & 56.2 \\
    \midrule
    \bf Ours & \bf 71.1 \\
    \bottomrule[1.3pt]
    \end{tabular}}
	\label{tab5}
\end{table}

\subsection{Results on Visual Grounding}
\subsubsection{Quantitative Performance Comparison}
We also evaluate the remote sensing visual grounding performance on DIOR-RSVG~\cite{dior-rsvg} to validate the effectiveness on region understanding and localization capabilities. TABLE~\ref{tab5} shows the experimental results along with those from existing VLMs, from general VLMs (\textit{i.e.,} Qwen-VL-Chat~\cite{qwen-vl} and CogVLM~\cite{cogvlm}) to remote sensing VLMs (\textit{i.e.,} GeoChat~\cite{geochat}, LHRS-Bot~\cite{lhrs-bot}, and VHM~\cite{vhm}), for the comparison. As shown in the table, the proposed framework outperforms the other baselines with a large performance margin. Such an experimental result corroborates that the effectiveness of multi-level semantic understanding, including fine-grained object-level contents.

\begin{table}[t!]
	\renewcommand{\arraystretch}{1.2}
	\renewcommand{\tabcolsep}{2.0mm}
    \centering
    \caption{The image captioning performance comparison on NWPU-Captions~\cite{nwpu-cap} and UCM-Captions~\cite{ucm}. The best and second runner-up performances are highlighted as \textbf{Bold text} and \underline{Underlined text}, respectively.}
    \resizebox{0.99\linewidth}{!}{
    \begin{tabular}{lcccc}
    \toprule[1.3pt]
    \multicolumn{5}{c}{\bf NWPU-Captions~\cite{nwpu-cap}} \\
    \cmidrule(lr){1-5}
    \bf Method & \bf BLEU-1 & \bf ROUGE & \bf METEOR & \bf BERT \\
    \midrule
    LLaVA-1.5~\cite{llava-1.5} & 7.2 & 12.1 & 19.8 & 84.9 \\
    Qwen2.5-VL~\cite{qwen2.5-vl} & 7.5 & 12.5 & 20.5 & 84.6 \\
    GeoChat~\cite{geochat} & 6.1 & 13.6 & 15.1 & 81.9 \\
    RSUniVLM~\cite{rsunivlm} & \bf 30.3 & 21.3 & 18.1 & 89.0 \\
    LHRS-Bot~\cite{lhrs-bot} & \underline{29.5} & \underline{26.8} & \underline{25.5} & \underline{89.3} \\
    \midrule
    \bf Ours & \underline{29.5} & \bf 27.4 & \bf 26.3 & \bf 89.5 \\
    \midrule\midrule
    \multicolumn{5}{c}{\bf UCM-Captions~\cite{ucm}} \\
    \cmidrule(lr){1-5}
    \bf Method & \bf BLEU-1 & \bf ROUGE & \bf METEOR & \bf BERT \\
    \midrule
    LLaVA-1.5~\cite{llava-1.5} & 7.4 & 12.9 & 21.6 & 85.0 \\
    Qwen2.5-VL~\cite{qwen2.5-vl} & 7.9 & 13.6 & 23.0 & 82.8 \\
    GeoChat~\cite{geochat} & \bf 43.7 & 10.9 & 14.9 & 81.5 \\
    RSUniVLM~\cite{rsunivlm} & 35.6 & 30.1 & \underline{28.5} & \underline{89.8} \\
    LHRS-Bot~\cite{lhrs-bot} & 35.0 & \underline{32.5} & 27.3 & 89.1 \\
    \midrule
    \bf Ours & \underline{42.3} & \bf 35.8 & \bf 30.6 & \bf 90.2 \\
    \bottomrule[1.3pt]
    \end{tabular}}
	\label{tab6}
\end{table}


\subsubsection{Qualitative Visualization Results}
For further validation, the qualitative visualization results of visual grounding task are illustrated in Fig.~\ref{fig6}. As shown in the figure, our proposed framework shows robust visual grounding performance across diverse scenes. Our framework properly perceives and localizes both large and small instances. For instance, as shown in the first and second examples, it successfully recognizes large ground track field and golf field. It also accurately captures small objects, such as vehicle and ship, as shown in the seventh and tenth examples.

\subsection{Results on Image Captioning}
We measure the performance on remote sensing image captioning task and compare the performance with existing baselines: LLaVA-1.5~\cite{llava-1.5}, Qwen2.5-VL~\cite{qwen2.5-vl}, GeoChat~\cite{geochat}, RSUniVLM~\cite{rsunivlm}, and LHRS-Bot~\cite{lhrs-bot}. TABLE~\ref{tab6} shows the performance on two remote sensing image captioning datasets: NWPU-Captions~\cite{nwpu-cap} and UCM-Captions~\cite{ucm}. We adopt BLEU-1, ROGUE-1, METEOR, and BERT F1 score as evaluation metrics for a comprehensive comparison on image captioning. As shown in the table, our framework consistently outperforms other model across most of the metrics, obtaining six best performances and two second runner-up performances out of eight comparisons. These results validate the effectiveness of the proposed method-a hierarchical semantic understanding design with semantic augmentation and semantic experts-on remote sensing image captioning.

\subsection{Exploration on Aggregation Tokens}
In this section, we explore the impact of the number of aggregation tokens, in order to determine the default number of the tokens. The aggregation tokens are designed to summarize the information from the user query tokens, retrieved semantic tokens, and multi-level visual feature tokens. The performances are evaluated on three benchmarks (\textit{i.e.,} NWPU~\cite{nwpu}, RSVQA-HR~\cite{rsvqa}, and UCM-Captions~\cite{ucm}) with $72$, $144$, and $288$ aggregation tokens. The exploration results are shown in TABLE~\ref{tab7}. For NWPU, VQA-HR, and UCM-Captions, the evaluation metrics are adopted as follows: accuracy averaged across scene categories, accuracy averaged across presence and comparison subsets, and BERT Score, respectively. As described in the table, $144$ aggregation tokens yields the most robust performance among the tested settings. Based on this observation, $144$ aggregation tokens are adopted as our default configuration.

\begin{table}[t!]
	\renewcommand{\arraystretch}{1.2}
	\renewcommand{\tabcolsep}{1.7mm}
    \centering
    \caption{Exploration for the number of aggregation tokens. The experiments are conducted on three benchmarks: NWPU (Scene Classification, Accuracy)~\cite{nwpu}, RSVQA-HR (VQA, Accuracy)~\cite{rsvqa}, and UCM-Captions (Image Captioning, BERT Score)~\cite{ucm}.}
    \resizebox{0.99\linewidth}{!}{
    \begin{tabular}{cccc}
    \toprule[1.3pt]
    \bf Number of Agg. Tokens & \bf NWPU & \bf RSVQA-HR & \bf UCM-Captions \\
    \midrule
    72 & 89.8 & 92.7 & 90.0 \\
    \bf 144 & \bf 90.4 & \bf 93.1 & \bf 90.2 \\
    288 & 89.9 & 93.0 & 90.0 \\
    \bottomrule[1.3pt]
    \end{tabular}}
	\label{tab7}
\end{table}

\section{Discussion}
\noindent
\textbf{Efficiency of Semantic-aware Expert Modeling.} As described in Section~\ref{expert}, we implement the semantic experts using a row-rank decomposition. Specifically, the $l$-th expert is formulated as $E^l(\cdot) = \boldsymbol{V}^l \cdot \boldsymbol{U}^l(\cdot)$, where $\boldsymbol{U}^l \in \mathbb{R}^{d_h \times d_r}$ and $\boldsymbol{V}^l \in \mathbb{R}^{d_r \times d_h}$, with $d_r \ll d_h$. In our implementation, the LLM hidden size $d_h$ is set to $3584$, and the reduced dimension $d_r$ is set to $512$, resulting in approximately $3.67$M parameters per semantic expert. In contrast, conventional Mixture-of-Experts (MoE) architectures usually require three projection matrices: a gating projection $\boldsymbol{P_G} \in \mathbb{R}^{d_h \times d_i}$, an up-projection $\boldsymbol{P_U} \in \mathbb{R}^{d_h \times d_i}$, and a down-projection $\boldsymbol{P_D} \in \mathbb{R}^{d_i \times d_h}$, where $d_h = 3584$ and $d_i = 18944$.Such a configuration leads to around $203.7$M parameters for each expert. It shows that our low-rank design requires only $1.8\%$ parameters used in standard MoE experts, enabling effective expert specialized processing across diverse semantic levels. \\
\noindent
\textbf{The Retrieval-based Semantic Augmentation Module.} A practical implementation consideration led to the design of a retrieval-based semantic augmentation module. As described in Section~\ref{retrieve}, we leveraged the captions from LHRS-Align-Recap~\cite{lhrs-bot-nova} to obtain rich and structured semantic descriptions covering scene- and instance-level semantics. However, we were unable to acquire most of the corresponding RS images from Google Earth API. Therefore, we sought to utilize such semantically rich descriptions--even in the absence of paired RS images, so that we devised a retrieval-based semantic augmentation approach. This architecture enables our framework to enhance semantic understanding of RS images without relying entirely on image-text pairs. As a result, our method not only addresses the data availability issue, but also shows the flexibility in incorporating textual semantics as an independent resource for downstream vision-language tasks. This approach may also be beneficial to other research domains, suffering from similar problems that cannot fully exploit one of the modalities, such as images. \\
\noindent
\textbf{Future Direction.} While our framework demonstrates promising performances across a wide range of remote sensing vision-language tasks, several open challenges and future research directions remain to be explored. First, although this work focuses on hierarchical semantic understanding in a given RS image, it could be extended to temporal analysis, such as change detection between multi-temporal remote sensing images. Second, it is worth to explore how to integrate additional modalities, such as synthetic aperture radar (SAR) images, into the framework for more comprehensive understanding. Lastly, even though our current framework supports horizontal bounding box (HBB) prediction, it could be also extended to predict oriented bounding box (OBB) for more accurate object localization in RS scenes in the future.
\section{Conclusion}
This paper proposed a novel Large Vision and Language Model (LVLM) framework designed for the remote sensing (RS) domain, which effectively addresses the inherent problem in RS imagery through multi-level semantic understanding. Our proposed framework consists of two core components: Semantic-augment Multi-level Alignment and Semantic-aware Expert Modeling. By leveraging the retrieval-based Semantic Augmentation Module, the framework effectively enriches visual prompts with contextual and hierarchical semantic information obtained from a curated RS semantic database. In addition, we devise Semantic-aware Expert Modeling with a mixture of multi-level semantic experts, enabling the framework to process and reason over multi-level semantic cues separately. Through extensive experiments on multiple RS vision-language tasks--scene classification, VQA, visual grounding, and image captioning, the proposed framework shows consistent effectiveness of our framework across a variety of tasks and benchmarks in both quantitative and qualitative manners. These experimental results demonstrate the importance of explicitly modeling hierarchical semantic understanding when adapting LVLMs to the RS domain. We expect this study to serve as a foundational step toward building next-generation RS LVLMs, and to inspire future research in developing more semantically-aware and domain-adapted multimodal models.

\bibliographystyle{IEEEtran}
\bibliography{IEEEabrv,main}

\end{document}